# Corollaries on the fixpoint completion: studying the stable semantics by means of the Clark completion

Pascal Hitzler[**]

Department of Computer Science, Dresden University of Technology
www.wv.inf.tu-dresden.de/∼pascal/
phitzler@inf.tu-dresden.de

**Abstract.** The fixpoint completion fix($P$) of a normal logic program $P$ is a program transformation such that the stable models of $P$ are exactly the models of the Clark completion of fix($P$). This is well-known and was studied by Dung and Kanchanasut [15]. The correspondence, however, goes much further: The Gelfond-Lifschitz operator of $P$ coincides with the immediate consequence operator of fix($P$), as shown by Wendt [51], and even carries over to standard operators used for characterizing the well-founded and the Kripke-Kleene semantics. We will apply this knowledge to the study of the stable semantics, and this will allow us to almost effortlessly derive new results concerning fixed-point and metric-based semantics, and neural-symbolic integration.

## 1 Introduction

The fixpoint completion of normal logic programs was introduced in [15], and independently under the notion of residual program in [9]. In essence, the fixpoint completion fix($P$) of a given program $P$ is obtained by performing a complete unfolding through all positive body literals in the program, and by disregarding all clauses with remaining positive body literals. Its importance lies in the fact that the stable models [20] of $P$ are exactly the supported models of fix($P$), i.e. the models of the Clark completion [11] of fix($P$). Also, the well-founded model [50] of $P$ is exactly the Fitting or Kripke-Kleene model [16] of fix($P$). These correspondences are well-known and have been employed by many authors for investigating the stable and the well-founded semantics, see e.g. [7].

The relation between a program and its fixpoint completion, however, is not exhausted by the correspondences between the different semantics just mentioned: It also concerns the semantic operators underlying these semantics, as shown in [51]. The virtue of this observation lies in the fact that it allows to carry over operator-based results on the supported, respectively, Fitting semantics, to the stable, respectively, well-founded semantics. To the best of our knowledge, this has not been noted before.

[**] This work was supported by a fellowship within the Postdoc-Programme of the German Academic Exchange Service (DAAD) and carried out while the author was visiting the Department of Electrical Engineering and Computer Science at Case Western Reserve University, Cleveland, Ohio.

In this paper, we display the strength of the operator-based correspondence by drawing a number of corollaries on the stable semantics from it. While these results are of interest in their own right, they do not constitute the main point we want to make here. Some of them are not even new, although we give new proofs. The goal of this paper is to provide a new technical tool for studying the stable and the well-founded semantics, namely the correspondences via the fixpoint completion between the semantic operators mentioned. To display this, we draw several corollaries from results in the literature, which are all valid for logic programs over a first-order language.

The structure of the paper is as follows. In Section 2 we recall the fixpoint completion and the results due to [51] which provide the starting points for our report. In Section 3 we study continuity of the Gelfond-Lifschitz operator in the Cantor topology, thereby providing technical results which will be of use later. In Section 4 we study methods for obtaining stable models by means of limits of iterates of the Gelfond-Lifschitz operator, and in Section 5 we will discuss results on the representation of logic programs by artificial neural networks. We briefly conclude in Section 6.

*Acknowledgement.* Thanks go to Matthias Wendt for helpful discussions and comments.

## 2 The Fixpoint Completion

A (*normal*) *logic program* is a finite set of universally quantified *clauses* of the form
$$\forall (A \leftarrow L_1 \wedge \cdots \wedge L_n),$$
where $n \in \mathbb{N}$ may differ for each clause, $A$ is an atom in a first order language $\mathcal{L}$ and $L_1, \ldots, L_n$ are literals, that is, atoms or negated atoms, in $\mathcal{L}$. As is customary in logic programming, we will write such a clause in the form
$$A \leftarrow L_1, \ldots, L_n,$$
in which the universal quantifier is understood, or even as
$$A\,\text{:-}\,L_1, \ldots, L_n$$
following Prolog notation. Then $A$ is called the *head* of the clause, each $L_i$ is called a *body literal* of the clause and their conjunction $L_1, \ldots, L_n$ is called the *body* of the clause. We allow $n = 0$, by an abuse of notation, which indicates that the body is empty; in this case the clause is called a *unit clause* or a *fact*. If no negation symbol occurs in a logic program, the program is called a *definite* logic program. The Herbrand base underlying a given program $P$, i.e. the set of all ground instances of atoms over $\mathcal{L}$, will be denoted by $B_P$, and the set of all

Herbrand interpretations by $I_P$, and we note that the latter can be identified simultaneously with the power set of $B_P$ and with the set $\mathbf{2}^{B_P}$ of all functions mapping $B_P$ into the set $\mathbf{2}$ consisting of two distinct elements. Since the set $I_P$ is the power set of $B_P$, it carries set-inclusion as natural ordering, which makes it a complete lattice. By $\mathsf{ground}(P)$ we denote the (possibly infinite) set of all ground instances of clauses in $P$.

The *single-step* or *immediate consequence operator* [37] of $P$ is defined as a function $T_P : I_P \to I_P$, where $T_I(I)$ is the set of all $A \in B_P$ for which there exists a clause $A \leftarrow L_1, \ldots, L_n$ with $I \models L_i$ for all $i = 1, \ldots, n$. A *supported model* of $P$ is a fixed point of $T_P$. Supported models correspond to models of the Clark completion of $P$, as noted in [1]. The pre-fixed points of $T_P$, i.e. interpretations $I \in I_P$ with $I \subseteq T_P(I)$, are exactly the Herbrand models of $P$, in the sense of first-order logic. If $P$ is definite, then $T_P$ is in fact a Scott- (or order-) continuous operator on $I_P$ [37], and its least fixed point $\mathsf{fix}(T_P)$ coincides with the least Herbrand model of $P$. The least fixed point, in this case, can be obtained as $\mathsf{fix}(T_P) = T_P \uparrow \omega := \sup_n (T_P \uparrow n) = \bigcup_n T_P \uparrow n$, where $T_P \uparrow 0 = \emptyset$ and recursively $T_P \uparrow (n+1) = T_P(T_P \uparrow n)$.

The *Gelfond-Lifschitz transformation* [20] of a program $P$ with respect to an interpretation $I$ is denoted by $P/I$, and consists of exactly those clauses $A \leftarrow A_1, \ldots, A_n$, where $A_1, \ldots, A_n \in B_P$, for which there exists a clause $A \leftarrow A_1, \ldots, A_n, \neg B_1, \ldots, \neg B_m$ in $\mathsf{ground}(P)$ with $B_1, \ldots, B_m \in I$. Thus $P/I$ is a definite program, and $\mathsf{fix}(T_{P/I})$ is well-defined. The *Gelfond-Lifschitz operator* [20] of $P$ is now defined by $\mathrm{GL}_P : I_P \to I_P : I \mapsto \mathsf{fix}(T_{P/I})$. We call $I \in I_P$ a *stable model* of $P$ if it is a fixed point of $\mathrm{GL}_P$.

**Definition 1.** *A quasi-interpretation[1] is a set of clauses of the form $A \leftarrow \neg B_1, \ldots, \neg B_m$, where $A$ and $B_i$ are ground atoms for all $i = 1, \ldots, m$. Given a normal logic program $P$ and a quasi-interpretation $Q$, we define $T'_P(Q)$ to be the quasi-interpretation consisting of the set of all clauses*

$$A \leftarrow \mathtt{body}_1, \ldots, \mathtt{body}_n, \neg B_1, \ldots, \neg B_m$$

*for which there exists a clause*

$$A \leftarrow A_1, \ldots, A_n, \neg B_1, \ldots, \neg B_m$$

*in $\mathsf{ground}(P)$ and clauses $A_i \leftarrow \mathtt{body}_i$ in $Q$ for all $i = 1, \ldots, n$. We explicitly allow the cases $n = 0$ or $m = 0$ in this definition.*

Note that the set of all quasi-interpretations is a complete partial order (cpo) with respect to set-inclusion. It was shown in [15], that for normal programs $P$, the operator $T'_P$ is Scott-continuous on the set of all quasi-interpretations. So we can define the *fixpoint completion* $\mathsf{fix}(P)$ of $P$ by $\mathsf{fix}(P) = T'_P \uparrow \omega$, i.e. $\mathsf{fix}(P)$ is the least fixed point of the operator $T'_P$.

The following was reported in [51].

---

[1] This notion is due to [15]. We stick to the old terminology, although quasi-interpretations should really be thought of as, and indeed are, programs with negative body literals only.

**Theorem 1.** *For any normal program $P$ and (two-valued) interpretation $I$, we have*
$$\mathrm{GL}_P(I) = T_{\mathsf{fix}(P)}(I).$$

*Proof.* We show first that for every $A \in \mathrm{GL}_P(I)$ there exists a clause in $\mathsf{fix}(P)$ with head $A$ whose body is true in $I$, which implies $A \in T_{\mathsf{fix}(P)}(I)$. We show this by induction on the powers of $T_{P/I}$; recall that $\mathrm{GL}_P(I) = T_{P/I}\!\uparrow\!\omega$.

For the base case $T_{P/I}\!\uparrow\!0 = \emptyset$ there is nothing to show.

So assume now that for all $A \in T_{P/I}\!\uparrow\!n$ there exists a clause in $\mathsf{fix}(P)$ with head $A$, whose body is true in $I$. For $A \in T_{P/I}\!\uparrow\!(n+1)$ there exists a clause $A \leftarrow A_1, \ldots, A_n$ in $P/I$ such that $A_1, \ldots, A_n \in T_{P/I}\!\uparrow\!n$, hence by construction of $P/I$ there is a clause $A \leftarrow A_1, \ldots, A_n, \neg B_1, \ldots, \neg B_m$ in $\mathsf{ground}(P)$ with $B_1, \ldots, B_m \notin I$. By induction hypothesis we obain that for each $i = 1, \ldots, n$ there exists a clause $A_i \leftarrow \mathsf{body}_i$ in $\mathsf{fix}(P)$ with $I \models \mathsf{body}_i$, hence $A_i \in T_{\mathsf{fix}(P)}(I)$. So by definition of $T'_P$ the clause $A \leftarrow \mathsf{body}_1, \ldots \mathsf{body}_n, \neg B_1, \ldots, \neg B_m$ is contained in $\mathsf{fix}(P)$. From $I \models \mathsf{body}_i$ and $B_1, \ldots, B_m \notin I$ we obtain $A \in T_{\mathsf{fix}(P)}(I)$ as desired.

This closes the induction argument and we obtain $\mathrm{GL}_P(I) \subseteq T_{\mathsf{fix}(P)}(I)$.

Now conversely, assume that $A \in T_{\mathsf{fix}(P)}(I)$. We show that $A \in \mathrm{GL}_P(I)$ by proving inductively on $k$ that $T_{T'_P\uparrow k}(I) \subseteq \mathrm{GL}_P(I)$ for all $k \in \mathbb{N}$.

For the base case, we have $T_{T'_P\uparrow 0}(I) = \emptyset$ so there is nothing to show.

So assume now that $T_{T'_P\uparrow k}(I) \subseteq \mathrm{GL}_P(I)$, and let $A \in T_{T'_P\uparrow(k+1)}(I) \setminus T_{T'_P\uparrow k}(I)$. Then there exists a clause $A \leftarrow \mathsf{body}_1, \ldots, \mathsf{body}_n, \neg B_1, \ldots, \neg B_m$ in $T'_P \uparrow (k+1)$ whose body is true in $I$. Thus $B_1, \ldots, B_m \notin I$ and for each $i = 1, \ldots, n$ there exists a clause $A_i \leftarrow \mathsf{body}_i$ in $T'_P \uparrow k$ with $\mathsf{body}_i$ true in $I$. So $A_i \in T_{T'_P\uparrow k}(I) \subseteq \mathrm{GL}_P(I)$. Furthermore, by definition of $T'_P$ there exists a clause $A \leftarrow A_1, \ldots, A_n, \neg B_1, \ldots, \neg B_m$ in $\mathsf{ground}(P)$, and since $B_1, \ldots, B_m \notin I$ we obtain $A \leftarrow A_1, \ldots, A_n \in P/I$. Since we know that $A_1, \ldots, A_n \in \mathrm{GL}_P(I)$ we obtain $A \in \mathrm{GL}_P(I)$, and hence $T_{T'_P\uparrow(k+1)}(I) \subseteq \mathrm{GL}_P(I)$. This closes the induction argument and we obtain $T_{\mathsf{fix}(P)}(I) \subseteq \mathrm{GL}_P(I)$. □

The proof of Theorem 1 is taken directly from [52], which appeared in compressed form as [51]. We have included it here for completeness of the exhibition and because the result is central for the rest of this paper. This correspondence can also be carried over to the Fitting/well-founded semantics. More precisely, the following was shown in [51], from which Theorem 1 is an easy Corollary.

**Theorem 2.** *For any normal program $P$ and any three-valued interpretation $I$ we have $\Psi_P(I) = \Phi_{\mathsf{fix}(P)}(I)$, where $\Psi_P$ is the operator due to [6] used for characterizing three-valued stable models of $P$, and $\Phi_{\mathsf{fix}(P)}$ is the operator from [16] used for characterizing the Fitting or Kripke-Kleene semantics of $\mathsf{fix}(P)$.*

We do not include details on this result here since we will need it only in passing in the sequel. The interested reader should consult [51]. A corollary from the result just mentioned is that the well-founded model of some given program $P$ coincides with the Fitting model of $\mathsf{fix}(P)$.

## 3  Continuity

Theorem 1 enables us to carry over results on the single-step operator, respectively on the supported-model semantics, to the Gelfond-Lifschitz operator respectively the stable-model semantics. The following observation is of technical importance.

**Proposition 1.** *Let $P$ be a definite program, $A \in B_P$, and $n \in \mathbb{N}$. Then $A \in T_P{\uparrow}n$ if and only if $A \leftarrow$ is a clause in $T'_P{\uparrow}n$.*

*Proof.* Let $A \in T_P \uparrow n$ for some $n \in \mathbb{N}$. We proceed by induction on $n$. If $n = 1$, then there is nothing to show. So assume that $n > 1$. Then there is a clause $A \leftarrow \texttt{body}$ in $\mathsf{ground}(P)$ such that all atoms $B_i$ in $\texttt{body}$ are contained in $T_P{\uparrow}(n-1)$, and by induction hypothesis there are claues $B_i \leftarrow$ in $T'_P{\uparrow}(n-1)$. Unfolding these clauses with $A \leftarrow \texttt{body}$ shows that $A \leftarrow$ is also contained in $T'_P{\uparrow}n$.

Conversely, assume there is a clause $A \leftarrow$ in $T'_P \uparrow n$. We proceed again by induction. If $n = 1$, there is nothing to show. So let $n > 1$. Then there exists a clause $A \leftarrow A_1, \ldots, A_k$ in $\mathsf{ground}(P)$ and clauses $A_i \leftarrow$ in $T'_P \uparrow (n-1)$. By induction hypothesis, we obtain $A_i \in T_P{\uparrow}(n-1)$ for all $i$, and hence $A \in T_P{\uparrow}n$. □

Since the single-step operator is not monotonic in general, several authors have made use of metric-based [17,18,22,25,26,27,29,46] or even topological [3,4,22,24,43,45,47] methods for obtaining fixed-points and hence supported models of the programs in question. Central to these investigations is the Cantor topology $Q$ on $I_P$, which was studied as the *query topology* in [4] and in more general terms as the *atomic topology* in [45]. It is the product topology on $\{\mathbf{t}, \mathbf{f}\}^{B_P}$, where the set of truth values $\{\mathbf{t}, \mathbf{f}\}$ is endowed with the discrete topology, and we refer to [53] for basic notions of topology. A subbase of the Cantor topology can be given as

$$\{\{I \in I_P \mid I \models L\} \mid L \text{ is a ground literal}\},$$

which was noted in [45]. We can now employ Theorem 1 to carry over some of these results to the treatment of the Gelfond-Lifschitz operator and the stable semantics.

Given a program $P$, we know by Theorem 1 that $\mathrm{GL}_P$ is continuous at some $I \in I_P$ in $Q$ if and only if $T_{\mathsf{fix}(P)}$ is continuous at $I$. This gives rise to the following theorem.

**Theorem 3.** *Let $P$ be a normal logic program and let $I \in I_P$. Then $\mathrm{GL}_P$ is continuous at $I$ in $Q$ if and only if whenever $\mathrm{GL}_P(I)(A) = \mathbf{f}$, then either there is no clause with head $A$ in $\mathsf{ground}(P)$ or there exists a finite set $S(I, A) = \{A_1, \ldots, A_k\} \subseteq B_P$ such that $I(A_i) = \mathbf{t}$ for all $i$ and for every clause $A \leftarrow \texttt{body}$ in $\mathsf{ground}(P)$ at least one $\neg A_i$ or some $B$ with $\mathrm{GL}_P(I)(B) = \mathbf{f}$ occurs in $\texttt{body}$.*

*Proof.* The proof is based on the characterization of continuity of the $T_P$-operator given in [45], in the formulation which can be found in [29, Theorem 45], which reads as follows.

> The single-step operator $T_P$ is continuous in $Q$ if and only if, for
> each $I \in I_P$ and for each $A \in B_P$ with $A \notin T_P(I)$, either there
> is no clause in $P$ with head $A$ or there is a finite set $S(I,A) = \{A_1, \ldots, A_k, B_1, \ldots, B_{k'}\}$ of elements of $B_P$ with the following properties:
> (i) $A_1, \ldots, A_k \in I$ and $B_1, \ldots, B_{k'} \notin I$.
> (ii) Given any clause $C$ with head $A$, at least one $\neg A_i$ or at least one $B_j$ occurs in the body of $C$.

Using this and Theorem 1, and by observing that there are no positive body atoms occuring in $\mathsf{fix}(P)$, we obtain the following:

> $\mathrm{GL}_P$ is continuous at $I$ if and only if whenever $\mathrm{GL}_P(I)(A) = \mathbf{f}$, then
> either there exists no clause with head $A$ in $\mathsf{fix}(P)$ or there exists a
> finite set $S(I,A) = \{A_1, \ldots, A_k\} \subseteq B_P$ such that $I(A_i) = \mathbf{t}$ for all $i$
> and for every clause $A \leftarrow \mathtt{body}$ in $\mathsf{fix}(P)$ at least one $\neg A_i$ occurs in
> $\mathtt{body}$.

So let $P$ be such that $\mathrm{GL}_P$ is continuous at $I$. If there is no clause with head $A$ in $\mathsf{ground}(P)$, then there is nothing to show. So assume that there is a clause with head $A$ in $\mathsf{ground}(P)$. We already know that then there exists a finite set $S(I,A) = \{A_1, \ldots, A_k\} \subseteq B_P$ such that $I(A_i) = \mathbf{t}$ for all $i$ and for every clause $A \leftarrow \mathtt{body}$ in $\mathsf{fix}(P)$ at least one $\neg A_i$ occurs in $\mathtt{body}$. Now let $A \leftarrow B_1, \ldots, B_k, \neg C_1, \ldots, \neg C_m$ be a clause in $\mathsf{ground}(P)$ and assume that no $\neg A_i$ occurs in its body. We show that there is some $B_i$ with $\mathrm{GL}_P(I)(B_i) = \mathbf{f}$. Assume the contrary, i.e. that $\mathrm{GL}_P(I)(B_i) = \mathbf{t}$ for all $i$. Then for each $B_i$ we have $B_i \in \mathrm{GL}_P(I) = T_{P/I}{\uparrow}\omega$. As in the proof of Proposition 1 we derive that there is a clause $A \leftarrow \neg D_1, \ldots, \neg D_n, \neg C_1, \ldots, \neg C_m$ in $\mathsf{fix}(P)$ with $D_j \notin I$ for all $j = 1, \ldots, n$. Since the clause $A \leftarrow \neg D_1, \ldots, \neg D_n, \neg C_1, \ldots, \neg C_m$ is contained in $\mathsf{fix}(P)$, we know that some atom from the set $S(I,A)$ must occur in its body. It cannot occur as any $D_i$ because $I(D_j) = \mathbf{f}$ for all $i$. It also cannot occur as any $C_i$ by assumption. So we obtain a contradiction, which finishes the argument.

Conversely, let $P$ be such that the condition on $\mathrm{GL}_P$ in the statement of the theorem holds. We will again make use of the observation made at the beginning of this proof. So let $A \in B_P$ with $\mathrm{GL}_P(I)(A) = \mathbf{f}$. If there is no clause with head $A$ in $\mathsf{fix}(P)$, then there is nothing to show. So assume there is a clause with head $A$ in $\mathsf{fix}(P)$. Then there is a clause with head $A$ in $P$, and by assumption we know that there exists a finite set $S(I,A) = \{A_1, \ldots, A_k\} \subseteq B_P$ such that $I(A_i) = \mathbf{t}$ for all $i$ and for every clause $A \leftarrow \mathtt{body}$ in $\mathsf{ground}(P)$ at least one $\neg A_i$ or some $B$ with $\mathrm{GL}_P(I)(B) = \mathbf{f}$ occurs in $\mathtt{body}$. Now let $A \leftarrow \neg B_1, \ldots, \neg B_n$ be a clause in $\mathsf{fix}(P) = T'_P{\uparrow}\omega$, i.e. there is $k \in \mathbb{N}$ with $A \leftarrow \neg B_1, \ldots, \neg B_n$ contained in $T'_P{\uparrow}k$. Note that $n = 0$ is impossible since this would imply $\mathrm{GL}_P(I)(A) = \mathbf{t}$ contradicting the assumption on $A$. We proceed by induction on $k$. If $k = 1$, then $A \leftarrow \neg B_1, \ldots, \neg B_n$ is contained in $\mathsf{ground}(P)$, hence one of the $B_j$ is contained in $S(I,A)$ which suffices. For $k > 1$, there is a clause $A \leftarrow C_1, \ldots, C_m, \neg D_1, \ldots, \neg D_{m'}$ in $\mathsf{ground}(P)$ and clauses $C_i \leftarrow \mathtt{body}_i$ in $T'_P{\uparrow}(k-1)$ which unfold to $A \leftarrow \neg B_1, \ldots, \neg B_n$. By assumption we either have $D_j \in S(I, A)$ for some $j$, in which case there remains nothing to show, or we have that $\mathrm{GL}_P(I)(C_i) = \mathbf{f}$ for some $i$. In the latter case we obtain that $\text{body}_i$ is non-empty by an argument similar to that of the proof of Proposition 1, so by assumption there is a (negated) atom in $\text{body}_i$, and hence in $\{B_1, \ldots, B_n\}$, which is also in $S(I, A)$, which finishes the proof. □

We can also observe the following special instance. A *local variable* is a variable occuring in some clause body but not in the corresponding head.

**Corollary 1.** *Let $P$ be a normal program without local variables. Then $\mathrm{GL}_P$ is continuous in $Q$.*

*Proof.* We employ Theorem 3. Let $I \in I_P$ and $A \in B_P$ with $\mathrm{GL}_P(I)(A) = \mathbf{f}$. Since $P$ has no local variables, it is of finite type. So the set $\mathcal{B}$ of all negated body atoms in clauses with head $A$ is finite. Let $S(I, A) = \{B \in \mathcal{B} \mid I(B) = \mathbf{f}\}$, which is also finite. If each clause with head $A$ contains some negated atom from $S(I, A)$, then there is nothing to show. So assume there is a clause $A \leftarrow A_1, \ldots, A_n, \neg B_1, \ldots, \neg B_m$ in $\text{ground}(P)$ with $B_j \notin S(I, A)$ for all $j$, i.e. $I(B_j) = \mathbf{t}$ for all $j$. But then $A \leftarrow A_1, \ldots, A_n$ is a clause in $P/I$ and $A \notin T_{P/I}\uparrow\omega$, which implies that there is some $i$ with $A_i \notin T_{P/I}\uparrow\omega = \mathrm{GL}_P(I)$, which finishes the argument by Theorem 3. □

Measurability is much simpler to deal with.

**Theorem 4.** *Let $P$ be a normal program. Then $\mathrm{GL}_P$ is measurable with respect to the $\sigma$-algebra $\sigma(Q)$ generated by $Q$.*

*Proof.* By [28, Theorem 2], which states that $T_P$ is measurable with respect to $\sigma(Q)$ for all $P$, we obtain that $T_{\text{fix}(P)}$ is measurable with respect to $\sigma(Q)$, and by Theorem 1 we know that $T_{\text{fix}(P)} \equiv \mathrm{GL}_P$. □

## 4 Obtaining models

As already mentioned above, topological methods in logic programming can for example be used for obtaining models of programs iteratively, although the underlying operator is not monotonic. The following variant of [29, Theorem 44] can be proven directly.

**Theorem 5.** *Let $P$ be a normal program and let $\mathrm{GL}_P$ be continuous and such that the sequence of iterates $\mathrm{GL}_P^m(I)$ converges in $Q$ to some $M \in I_P$. Then $M$ is a stable model of $P$.*

*Proof.* By continuity we obtain

$$M = \lim \mathrm{GL}_P^m(I) = \mathrm{GL}_P(\lim \mathrm{GL}_P^m(I)) = \mathrm{GL}_P(M).$$

□

We can also employ knowledge about relationships between the single-step operator and the Fitting operator [16]. The latter is defined on three-valued interpretations, which consist of sets of ground *literals* (instead of ground *atoms*) which do not contain complementary literals. As such, they carry set-inclusion as an ordering, which renders the space $I_{P,3}$ of all three-valued interpretations a complete partial order (cpo). It is in fact exactly the Plotkin domain $\mathbb{T}^\omega$ due to [41]. Alternatively, we can understand three-valued interpretations as mappings from atoms to the set $\{\mathbf{f}, \mathbf{u}, \mathbf{t}\}$ of truth values, where $\mathbf{u}$ stands for *undefined* or *undetermined*. The Fitting operator $\Phi_P$, for given program $P$, is now defined as a function $\Phi_P : I_{P,3} \to I_{P,3} : I \mapsto t_P(I) \cup f_P(I)$, where $t_P(I)$ contains all $A \in B_P$ for which there exists a clause $A \leftarrow L_1, \ldots, L_n$ in ground$(P)$ with $L_1, \ldots, L_n \in I$, and $f_P(I)$ contains all $\neg A$ such that for all clauses $A \leftarrow L_1, \ldots, L_n$ in ground$(P)$ there is at least one $L_i \notin I$. It was shown in [16] that $\Phi_P$ is a monotonic operator on $I_{P,3}$.

If $I$ is a three-valued interpretation, then $I^+$ denotes the two-valued interpretation assigning truth value $\mathbf{t}$ to exactly those atoms which are true in $I$.

**Proposition 2.** *Let $P$ be a normal program and assume that the well-founded model $M$ of $P$ is total (i.e. every atom is true or false in it). Then $\mathrm{GL}_P^n(\emptyset)$ converges in $Q$ to $M^+$, and $M^+$ is the unique stable model of $P$.*

*Proof.* This follows immediately from Theorem 1 and [24, Theorem 4.4], which shows the following.

> If $M = \Phi_R {\uparrow} \omega$ is total, then $T_R^n(\emptyset)$ converges in $Q$ to $M^+$, and $M^+$ is the unique supported model of $R$.

□

Metric-based approaches also carry over. A *level mapping* is a mapping from $B_P$ to some ordinal $\alpha$. A program $P$ is *locally stratified* [44] if there exists a level mapping $l : B_P \to \alpha$, where $\alpha$ is some ordinal, such that for each clause $A \leftarrow A_1, \ldots, A_m, \neg B_1, \ldots, \neg B_n$ in ground$(P)$ we have $l(A) \geq l(A_i)$ and $l(A) > l(B_j)$ for all $i$ and $j$. It is called *locally hierarchical* [10], if additionally $l(A) > l(A_i)$ for all $i$. Given a level mapping $l : B_P \to \alpha$, we denote by $\Gamma_l$ the set of all symbols $2^{-\beta}$ for $\beta \leq \alpha$, ordered by $2^{-\beta} < 2^{-\gamma}$ iff $\gamma < \beta$. $\Gamma_l$ can be understood as a subset of the reals if $\alpha = \omega$, i.e. if $l$ maps into the natural numbers. For two (two-valued) interpretations $I$ and $J$, we define $d_l(I, J) = 2^{-\beta}$, where $\beta$ is the least ordinal such that there is an atom of level $\beta$ on which $I$ and $J$ disagree. If $\alpha = \omega$, then $d_l$ is an ultrametric on $I_P$, and this construction was put to use e.g. in [17]. In the general case, $d_l$ is a generalized ultrametric on $I_P$, as used in logic programming e.g. in [25,29,43]. A mapping $f$ is called *strictly contracting* with respect to a generalized ultrametric $d$ if $d(f(x), f(y)) < d(x, y)$ for all $x, y$ with $x \neq y$. Strictly contracting mappings have unique fixed points if the underlying generalized ultrametric space satisfies a completeness condition called *spherical completeness* [43].

**Theorem 6.** *Let $P$ be locally stratified with corresponding level mapping $l$. Then $\mathrm{GL}_P$ is strictly contracting with respect to $d_l$, which is spherically complete. If $l*

*maps to* $\omega$, *then* $\mathrm{GL}_P$ *is a contraction with respect to* $d_l$. *Furthermore, in both cases,* $\mathrm{GL}_P$ *has a unique fixed point and* $P$ *has a unique stable model.*

*Proof.* If $P$ is locally stratified with respect to $l$, then $\mathsf{fix}(P)$ is locally hierarchical with respect to $l$. It thus suffices to apply Theorem 1 in conjunction with Theorem [47, Theorem 3.8], which shows the following.

> Let $R$ be a normal logic program which is locally hierarchical with respect to a level mapping $l : B_R \to \gamma$. Then $T_R$ is strictly contracting with respect to the generalized ultrametric $d_l$ induced by $l$. Therefore, $T_R$ has a unique fixed point and hence $R$ has a unique supported model.

□

With the remarks already made on the fact that the well-founded model of some given program $P$ coincides with the Fitting model of $\mathsf{fix}(P)$, for any normal program $P$, we can also derive the following result. *Dislocated generalized ultrametric spaces* are defined by relaxing one of the defining conditions on generalized ultrametrics, for details see [29]. Strictly contracting mappings can be defined analogously, and have similar properties.

**Theorem 7.** *Let $P$ be a program with total well-founded model $I \cup \neg(B_P \setminus I)$, with $I \subseteq B_P$. Then $\mathrm{GL}_P$ is strictly contracting on the spherically complete dislocated generalized ultrametric space $(I_P, \varrho)$, where we have $\varrho(J, K) = \max\{d_l(J, I), d_l(I, K)\}$ for all $J, K \in I_P$, and $l$ is defined by $l(A)$ to be the minimal $\alpha$ such that $\Phi_{\mathsf{fix}(P)}\!\uparrow\!(\alpha + 1)(A) = I(A)$.*

*Proof.* The program $P$ has a total well-founded model, which implies that $\mathsf{fix}(P)$ has a total Fitting model. So $l$ as given by the statement is well-defined, and $\mathsf{fix}(P)$ is $\Phi$-*accessible* in the sense of [29]. Now apply [29, Proposition 41], which shows that $T_P$ is strictly contracting for every $\Phi$-accessible program. □

## 5 Neural-symbolic integration

Intelligent systems based on logic programming on the one hand, and on artificial neural networks (sometimes called connectionist sytems) on the other, differ substantially. Logic programs are highly recursive and well understood from the perspective of knowledge representation: The underlying language is that of first-order logic, which is symbolic in nature and makes it easy to encode problem specifications directly as programs. The success of artificial neural networks lies in the fact that they can be trained using raw data, and in some problem domains the generalization from the raw data made during the learning process turns out to be highly adequate for the problem at hand, even if the training data contains some noise. Successful architectures, however, often do not use recursive (or recurrent) structures. Furthermore, the knowledge encoded by a trained neural network is only very implicitly represented, and no satisfactory methods for extracting this knowledge in symbolic form are currently known.

It would be very desirable to combine the robust neural networking machinery with symbolic knowledge representation and reasoning paradigms like logic programming in such a way that the strenghts of either paradigm will be retained. Current state-of-the-art research, however, fails by far to achieve this ultimate goal. As one of the main obstacles to be overcome we perceive the question how symbolic knowledge can be encoded by artificial neural networks: Satisfactory answers to this will naturally lead the way to knowledge extraction algorithms and to hybrid neural-symbolic systems.

Earlier attempts to integrate logic and connectionist systems have mainly been restricted to propositional logic, or to first-order logic without function symbols. They go back to the pioneering work by McCulloch and Pitts [39], and have led to a number of systems developed in the 80s and 90s, including Towell and Shavlik's KBANN [49], Shastri's SHRUTI [48], the work by Pinkas [40], Hölldobler [30], and d'Avila Garcez et al. [12,14], to mention a few, and we refer to [8,13,21] for comprehensive literature overviews.

Without the restriction to the finite case (including propositional logic and first-order logic without function symbols), the task becomes much harder due to the fact that the underlying language is infinite but shall be encoded using networks with a finite number of nodes. The sole approach known to us for overcoming this problem (apart from work on recursive autoassociative memory, RAAM, initiated by Pollack [42], which concerns the learning of recursive terms over a first-order language) is based on a proposal by Hölldobler et al. [32], spelled out first for the propositional case in [31], and reported also in [23]. It is based on the idea that logic programs can be represented — at least up to subsumption equivalence [38] — by their associated single-step operators. Such an operator can then be mapped to a function on the real numbers, which can under certain conditions in turn be encoded or approximated e.g. by feedforward networks with sigmoidal activation functions using an approximation theorem due to Funahashi [19].

We will carry over this result to the Gelfond-Lifschitz operator and the stable model semantics. Since the topology $Q$ introduced earlier is homeomorphic to the Cantor topology on the real line [45], there exists a homeomorphism $\iota : I_P \to \mathcal{C}$, where $\mathcal{C}$ is the Cantor set within the unit interval, endowed with the subspace topology inherited from the reals. We can thus embed any function $f : I_P \to I_P$ which is continuous in $Q$ as a continuous function $\iota(f) : \mathcal{C} \to \mathcal{C} : \iota(f)(x) = \iota(f(\iota^{-1}(x)))$. By well-known results, e.g. [19] as mentioned earlier, such functions can be approximated uniformly by artificial neural networks in many different network architectures.

**Theorem 8.** *Let $P$ be a normal logic program. Then $\mathrm{GL}_P$ can be approximated almost everywhere up to an arbitrarily chosen error bound by input-output functions of three-layer feedforward neural networks with sigmoidal activation functions. If $\mathrm{GL}_P$ is furthermore continuous in $Q$, then uniform approximation is possible on all of $\mathcal{C}$.*

*Proof.* We use Theorem 1. The first statement then follows from Theorem 4 together with a result from [33] saying that each measurable function can be approximated almost everywhere by three-layer feedforward networks in the indicated way — see also [28, Theorem 7]. The second statement follows from [19] or from [28, Theorem 5]. □

The references mentioned in the proof of Theorem 8 provide further results, in particular on error bounds, and they can also be carried over straightforwardly.

Another improvement on the basic results by Hölldobler et al [32] employed an alternative network architecture. In [2], results were provided for encoding and approximating $\iota(T_P)$ by iterated function systems, which in turn could be encoded using a recurrent neural networks structure. The advantage of this approach is that algorithms for constructing approximating networks can be given explicitly, in contrast to the results in [23,28,32]. These results also hinge on continuity or Lipschitz-continuity of $\iota(T_P)$ with respect to the Cantor topology only, and can be carried over to the Gelfond-Lifschitz operator in a straightforward way. The paper [5] provides related results using cellular automata, treating logic programs without local variables — a property which also carries over to the fixpoint completion. Hence these results carry over *mutatis mutandis* to the Gelfond-Lifschitz operator.

## 6 Conclusions

We have displayed the usefulness of the results reported in [51] to the operator-based analysis of knowledge representation under the stable semantics. We have shown that many results from the study of the supported-model semantics by means of the single-step operator can be carried over to the stable semantics almost without effort.

Our results are of a theoretical nature, and we do not propose to study them for implementation purposes. The idea to use the fixpoint completion for obtaining stable models (or similar constructions for obtaining answer sets or well-founded models etc.) of programs is already folklore knowledge in the community, and need not be further mentioned. The emphasis of our exhibition is on the observation that not only *models*, but also corresponding *semantic operators* are related by means of the fixpoint completion, and on the aspects which this new insight allows to study.

Our observations are valid for first-order languages including function symbols, a syntax whose study is often neglected in the non-monotonic reasoning community. It is not at all surprising, that for *finite* languages alternative methods of program transformation can be found, which allow for efficient computation of stable models [34,35,36].

---

[2] Available online as [52].